\documentclass{article}

\usepackage[preprint]{neurips_2026}

\usepackage[utf8]{inputenc}
\usepackage[T1]{fontenc}
\usepackage{hyperref}
\usepackage{url}
\usepackage{booktabs}
\usepackage{multirow}
\usepackage{amsfonts}
\usepackage{nicefrac}
\usepackage{microtype}
\usepackage{graphicx}
\usepackage{xcolor}
\usepackage{amsmath}
\usepackage{amssymb}
\usepackage{subcaption}
\usepackage{tikz}
\usetikzlibrary{arrows.meta, positioning, calc, decorations.markings, decorations.pathreplacing}
\usepackage{float}  
\usepackage{placeins}  

\usepackage{caption}
\title{SMolLM: Small Language Models Learn Small Molecular Grammar}

\author{%
  Akhil Jindal \\
  Moku \\
  \texttt{akhil@moku.gg} \\
  \And
  Harang Ju \\
  Carey Business School \\
  Johns Hopkins University \\
  \texttt{harang@jhu.edu} \\
}

\begin{document}

\maketitle

\begin{abstract}
Language models for molecular design have scaled to hundreds of millions of parameters, yet how they learn chemical grammar is poorly understood. We train SMolLM, a 53K-parameter weight-shared transformer, to generate novel SMILES with 95\% validity on the ZINC-250K drug-like-molecule benchmark, outperforming a standard GPT with 10 times more parameters. Mechanistically, the same block resolves SMILES constraints across passes in a fixed hierarchy: brackets first, rings second, and valence last, as shown by error classification and linear probing, with ablation isolating the bracket-matching head. Together, these results yield a compact, mechanistically interpretable molecular generator and a testbed for studying iterative computation in formal-language domains.
\end{abstract}

\section{Introduction}

Designing new molecules is a core task in drug discovery and materials science, and language models have become a major tool for molecular generation. From RNNs \citep{segler2018generating, olivecrona2017molecular} to transformer decoders \citep{bagal2022molgpt, ross2024gpmolformer, xu2026scaling}, these generators have grown by orders of magnitude in parameter count. However, whether such scale is necessary and what computations underlie valid generation remain poorly understood.

One promising way to reduce parameter count without reducing computational depth is to reuse the same transformer block repeatedly. Weight-shared transformers instantiate this idea by applying a single block iteratively for K passes. This matches the computational depth of a K-layer unshared stack with one block's worth of parameters. Such architectures are Turing-complete \citep{giannou2023looped}, can run iterative algorithms like gradient descent \citep{vonoswald2023gd}, and match or exceed standard transformers on algorithmic reasoning \citep{dehghani2019universal}, language modeling \citep{bae2025relaxed, reid2021subformer}, and vision \citep{goyal2026elt}. This makes them particularly suited to formal-language domains like SMILES, where validity depends on repeated local and long-range checks.

Weight sharing also makes generation interpretable by attributing behavior to computation rather than parameters. Prior work in mechanistic interpretability has decomposed circuits in standard transformers \citep{elhage2021mathematical, olsson2022induction} and in molecular transformers \citep{varadi2025circuits}. In these architectures, however, each layer has its own weights, so functional differences between layers can reflect differences in parameters rather than differences in computation.

Here, we show that a single 53K-parameter weight-shared transformer block, trained on ZINC-250K \citep{irwin2012zinc} SMILES strings (Simplified Molecular Input Line Entry System), generates novel SMILES with 95\% validity. At 53K parameters, the model is over an order of magnitude smaller than the smallest prior transformer-based molecular generator we compare against \citep{bagal2022molgpt, ross2024gpmolformer, xu2026scaling}, and outperforms an identically-trained 527K-parameter unshared GPT in validity (95.3\% vs.\ 87.6\%) using 10 times fewer parameters.

Beyond generating valid molecules, the block is mechanistically interpretable. With weight sharing, the same parameters run at every pass, so functional differences across passes reflect changes in computation, not changes in weights. SMILES validity depends on three hard constraints: brackets must match, ring digits must pair at long range, and atom valences must be chemically valid. Because each constraint has an exact symbolic check, we can measure how generation improves at every intermediate pass.

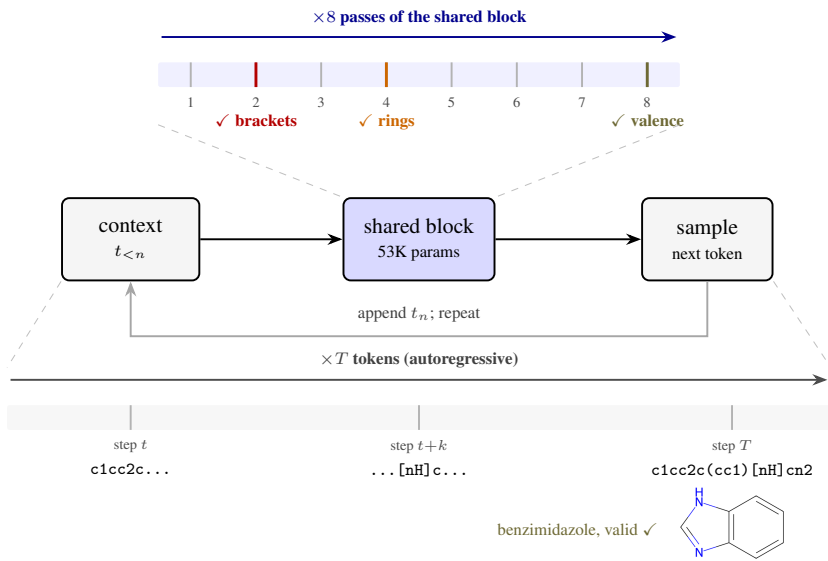
\begin{figure}[!t]
\centering
\resizebox{0.78\textwidth}{!}{%
\begin{tikzpicture}[
  font=\small,
  arr/.style={-{Stealth[length=2.2mm]}, thick},
  box/.style={draw, rounded corners=3pt, thick, align=center, minimum height=1.2cm},
  tok/.style={font=\ttfamily\scriptsize},
]
  \def\gxl{-3.8}
  \def\gxr{3.8}
  \def\gy{2.4}
  \def\gh{0.18}

  \fill[blue!6, rounded corners=1pt] (\gxl, \gy-\gh) rectangle (\gxr, \gy+\gh);

  \foreach \i in {1,...,8} {
    \pgfmathsetmacro\x{\gxl + (\i-0.5)*(\gxr-\gxl)/8}
    \draw[gray!60, thick] (\x, \gy-\gh) -- (\x, \gy+\gh);
    \node[font=\tiny, text=gray!60!black, anchor=north] at (\x, \gy-\gh-0.03) {\i};
  }
  \draw[arr, blue!55!black, thick] (\gxl+0.05, \gy+0.55) -- (\gxr-0.05, \gy+0.55);
  \node[font=\scriptsize\bfseries, text=blue!55!black, anchor=south]
        at ({(\gxl+\gxr)/2}, \gy+0.6) {$\times 8$ passes of the shared block};

  \pgfmathsetmacro\xA{\gxl + 1.5*(\gxr-\gxl)/8}
  \pgfmathsetmacro\xB{\gxl + 3.5*(\gxr-\gxl)/8}
  \pgfmathsetmacro\xC{\gxl + 7.5*(\gxr-\gxl)/8}
  \draw[red!70!black, line width=1.2pt] (\xA, \gy-\gh) -- (\xA, \gy+\gh);
  \draw[orange!80!black, line width=1.2pt] (\xB, \gy-\gh) -- (\xB, \gy+\gh);
  \draw[olive!60!black, line width=1.2pt] (\xC, \gy-\gh) -- (\xC, \gy+\gh);
  \node[font=\scriptsize\bfseries, text=red!70!black, anchor=north] at (\xA, \gy-\gh-0.28) {$\checkmark$ brackets};
  \node[font=\scriptsize\bfseries, text=orange!80!black, anchor=north] at (\xB, \gy-\gh-0.28) {$\checkmark$ rings};
  \node[font=\scriptsize\bfseries, text=olive!60!black, anchor=north] at (\xC, \gy-\gh-0.28) {$\checkmark$ valence};

  \node[box, fill=gray!8, minimum width=2.0cm] (ctx) at (-4.2, 0)
        {context\\\scriptsize $t_{<n}$};
  \node[box, fill=blue!15, minimum width=2.2cm] (blk) at (0, 0)
        {shared block\\\scriptsize 53K params};
  \node[box, fill=gray!8, minimum width=1.9cm] (smp) at (4.2, 0)
        {sample\\\scriptsize next token};

  \draw[arr] (ctx) -- (blk);
  \draw[arr] (blk) -- (smp);

  \draw[gray!50, dashed, thin] (\gxl, \gy-\gh-0.55) -- (blk.north west);
  \draw[gray!50, dashed, thin] (\gxr, \gy-\gh-0.55) -- (blk.north east);

  \draw[arr, gray!70, thick]
        (smp.south) -- ++(0, -0.8) -| (ctx.south);
  \node[font=\scriptsize, text=gray!50!black, anchor=north] at (0, -0.85) {append $t_n$; repeat};

  \def\txl{-6.0}
  \def\txr{6.0}
  \def\ty{-2.6}
  \def\th{0.18}

  \fill[gray!6, rounded corners=1pt] (\txl, \ty-\th) rectangle (\txr, \ty+\th);

  \draw[arr, gray!60!black, thick] (\txl+0.05, \ty+0.55) -- (\txr-0.05, \ty+0.55);
  \node[font=\scriptsize\bfseries, text=gray!50!black, anchor=south]
        at ({(\txl+\txr)/2}, \ty+0.6) {$\times T$ tokens (autoregressive)};

  \pgfmathsetmacro\xT{\txl + 0.15*(\txr-\txl)}
  \pgfmathsetmacro\xK{\txl + 0.50*(\txr-\txl)}
  \pgfmathsetmacro\xF{\txl + 0.88*(\txr-\txl)}

  \foreach \x/\lbl in {\xT/{step $t$}, \xK/{step $t{+}k$}, \xF/{step $T$}} {
    \draw[gray!60, thick] (\x, \ty-\th) -- (\x, \ty+\th);
    \node[font=\tiny, text=gray!60!black, anchor=north] at (\x, \ty-\th-0.03) {\lbl};
  }

  \node[tok, anchor=north] at (\xT, \ty-0.55) {c1cc2c...};

  \node[tok, anchor=north] at (\xK, \ty-0.55) {...[nH]c...};

  \node[tok, anchor=north] at (\xF, \ty-0.55) {c1cc2c(cc1)[nH]cn2};
  \node[anchor=north] at (\xF, \ty-0.82) {\includegraphics[width=1.65cm]{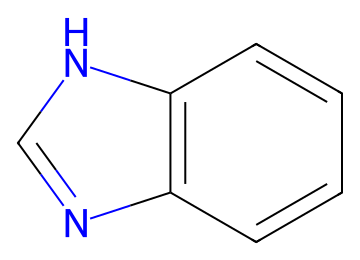}};
  \node[font=\scriptsize, text=olive!60!black, anchor=east] at (\xF-0.95, \ty-1.645) {benzimidazole, valid $\checkmark$};

  \draw[gray!50, dashed, thin] (ctx.south west) -- (\txl, \ty+\th+0.55);
  \draw[gray!50, dashed, thin] (smp.south east) -- (\txr, \ty+\th+0.55);
\end{tikzpicture}%
}
\caption{\textbf{Overview of SMolLM.} \emph{Top track:} for each emitted token, the same 53K shared block runs eight passes. By truncating inference at intermediate passes, we find that the shared block solves grammar in stages, with each stage retained as depth increases: brackets by pass 2, rings by pass 4, valence by pass 8. \emph{Bottom track:} the model autoregressively emits $T$ SMILES tokens. The molecule at step $T$ is benzimidazole, which is generated by WS-53K-s42 in a 10{,}000-sample run and absent from the training data.}
\label{fig:overview}
\end{figure}

Using three analyses (error classification across passes, a combined probing/sparse-autoencoder study of internal representations, and ablation), we show that the model learns chemical grammar pass by pass: brackets are resolved by pass 2, rings by pass 4, and valence by pass 8 (Fig.~\ref{fig:overview}). Sparse autoencoders and linear probing recover the same ordering in the internal representations, from individual tokens in early passes to bracket depth and ring state in middle passes (ring-state accuracy climbs from 79\% at pass 1 to 98\% by pass 5). Ablation localizes bracket matching to a single attention head in WS-206K, where ablating it at pass 1 raises bracket errors by 23 percentage points without changing ring or valence errors, while the smaller WS-53K distributes the same function across heads.

Together, these results show a shared block resolving constraints from local to global scope, suggesting that similar pass-wise structure may appear in other formal-language domains with hierarchical constraints (Section~\ref{sec:mechanism}). The weight-shared block thus serves as both a compact generator for molecular design and a small testbed for iterative computation.

\section{Experimental setup}

This section defines the SMILES generation task, the weight-shared and baseline models, and the training and evaluation protocol.

\paragraph{SMILES.}
SMILES is a text encoding of molecular graphs. For example, atoms can be written with symbols such as \texttt{C}, \texttt{N}, and \texttt{O}. Bonds can be implicit or written explicitly with symbols such as \texttt{=} and \texttt{\#}. Branches use parentheses, and ring closures use paired digits. The string \texttt{CC(=O)O}, for instance, encodes acetic acid. A generated string is valid only if brackets balance, ring digits pair, and atom valences are chemically possible. Validity rate is the fraction of generated strings that decode to real molecules via RDKit. We use SMILES rather than SELFIES \citep{krenn2020selfies} because SMILES leaves these grammar and chemistry constraints for the model to learn, whereas SELFIES guarantees validity by construction.

\paragraph{Architecture.}
All models are autoregressive transformers based on GPT \citep{radford2019language}. We compare three variants. Standard GPT uses learned positional embeddings and Gaussian Error Linear Unit (GELU) activations with a feed-forward expansion factor of 4. Modern GPT replaces these with rotary position embeddings (RoPE; \citealp{su2024roformer}) and Swish-Gated Linear Units (SwiGLU; \citealp{shazeer2020glu}). Weight-Shared (WS) reuses a single Modern GPT block for $D$ forward passes, equivalent to a looped transformer \citep{giannou2023looped}. We call $D$ its \emph{virtual depth}. All transformer variants apply pre-norm LayerNorm (LayerNorm before each sublayer), tie input/output embeddings, use dropout 0.1, and have no bias terms. We additionally evaluate Gated Recurrent Unit (GRU) baselines at parameter counts matched to WS-53K and WS-206K. Model names combine class and parameter count, so WS-53K denotes a weight-shared block with 53K parameters and default virtual depth $D{=}8$. GPT-527K and GPT-527K-mod denote unshared Standard and Modern GPTs, and GRU-53K denotes the GRU baseline. When a WS model uses a virtual depth other than the default $D{=}8$, its name includes a suffix such as \texttt{-D4} or \texttt{-D16}. WS-53K and WS-206K are the two weight-shared blocks that recur, with WS-206K depth variants for the virtual-depth sweep. Full configurations are in Appendix~\ref{app:configs}.

\paragraph{Training.}
We train on ZINC-250K \citep{irwin2012zinc}, the standard benchmark of 249{,}455 drug-like molecules for SMILES-based molecular generation \citep{bagal2022molgpt, ross2024gpmolformer}, with SMILES augmentation at $n_\text{aug}{=}10$ \citep{bjerrum2017smiles} and a character-level tokenizer (vocab 50, max length 128). We optimize with AdamW for 100 epochs at batch size 256, using a cosine learning-rate schedule (peak $5 \times 10^{-4}$, 200-step warmup), mixed-precision training (AMP), and gradient clipping at 1.0. Core experiments use 3 seeds (42, 43, 44); supporting ablations use 1 seed. The reported experiments used approximately 500 GPU-hours on AMD Instinct MI210/MI250 GPUs. Distillation and DPO settings are given in Appendix~\ref{app:kd_ablation}.

\paragraph{Evaluation.}
We sample $N{=}1{,}000$ molecules per model with temperature $0.8$ and top-$k{=}40$. We report RDKit validity, Fréchet ChemNet Distance (FCD; \citealp{preuer2018fcd}), internal diversity ($1$ minus mean pairwise Tanimoto similarity, a Jaccard-style overlap measure, computed on Morgan radius-2 1024-bit fingerprints), uniqueness, and novelty, defined as the fraction of generated molecules absent from the ZINC-250K training set. We categorize invalid SMILES by failure mode for the error classification analysis (Section~\ref{sec:error}). Further details are in Appendix~\ref{app:setup}.

\section{Compact molecular generation}
\label{sec:compression}

A 53K-parameter weight-shared transformer reaches 95.3\% SMILES validity by sharing a single block across 8 passes, outperforming an unshared GPT with 10 times more parameters and approaching the validity of a GPT with 60 times more parameters.

\subsection{Pareto frontier}

We first ask whether weight sharing improves the tradeoff between parameter count and SMILES validity. We show that weight-shared models dominate the sub-megaparameter Pareto frontier. At every budget below 1M, the weight-shared curve delivers higher validity at fewer parameters (Fig.~\ref{fig:scaling}; full numbers in Appendix~\ref{app:configs}, Table~\ref{tab:pareto}). WS-53K outperforms GPT-527K by 7.7 points (95.3\% vs.\ 87.6\%) and matches its internal diversity (0.852 vs.\ 0.851), at 10 times fewer parameters. The generated molecules show little evidence of memorization: across 3 seeds, 0.13\% of valid WS-53K samples are exact training copies, and the mean nearest-neighbor Tanimoto similarity to training molecules is 0.52, below the nearest-neighbor similarity within the training set (0.59). WS-53K also matches GPT-3.2M to within 5 points (95.3\% vs.\ 99.4\%) at 60 times fewer parameters. FCD, a distributional-similarity metric where lower is better, is modestly higher for weight-shared models (2.5--2.9 vs.\ 2.0--2.4), trading some distributional fidelity for parameter efficiency. Thus, weight sharing shifts the compact-generation frontier rather than merely matching smaller baselines.

\begin{figure}[t]
\centering
\includegraphics[width=0.53\textwidth]{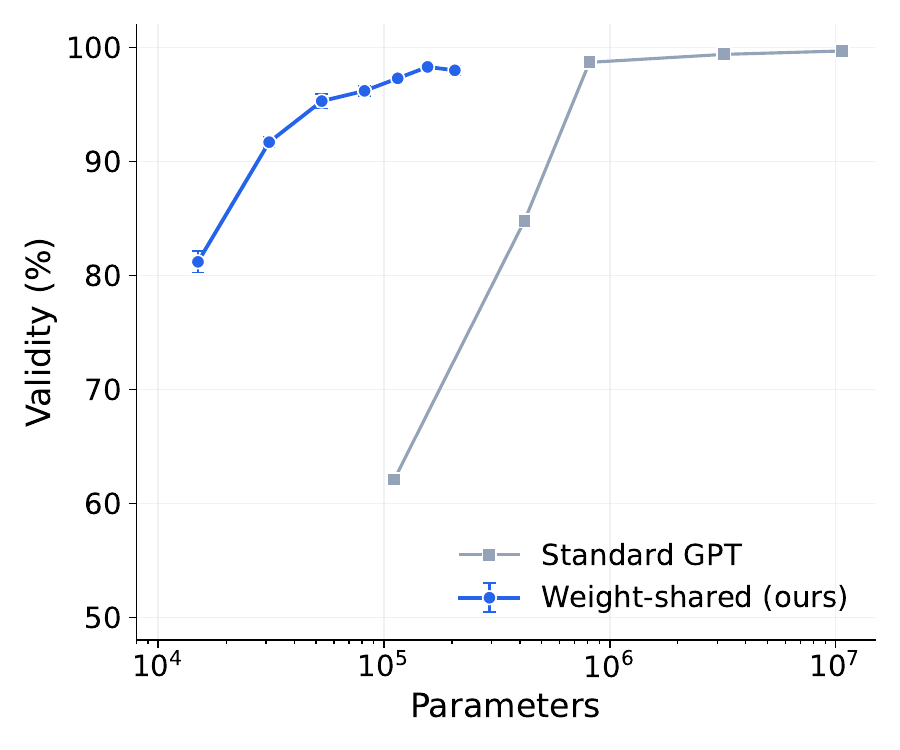}
\caption{Pareto frontier. Weight-shared models dominate unshared GPTs below 1M parameters.}
\label{fig:scaling}
\end{figure}

\subsection{Virtual depth}

We vary how many times the shared block is applied, measuring whether validity improves over repeated passes. Validity rises sharply with virtual depth and plateaus at $D{=}8$: 33\% $\rightarrow$ 91\% $\rightarrow$ 97\% $\rightarrow$ 99\% $\rightarrow$ 99\% across $D \in \{1, 2, 4, 8, 16\}$ (WS-206K, 3 seeds per cell). The $D{=}1$ value (33.0~$\pm$~2.3\%, 3 seeds, 1{,}000 samples) matches the separately-trained WS-206K-D1 row of Table~\ref{tab:errors} (33.0~$\pm$~2.3\%). The 58-point rise from $D{=}1$ to $D{=}2$, followed by diminishing returns through $D{=}8$ and a plateau at $D{=}16$, suggests that repeated passes iteratively refine validity.

We next isolate whether the gain comes from iteration itself rather than from other block-level architectural choices. An unshared Standard GPT baseline with four attention heads reaches 91.6~$\pm$~0.3\% validity. The matched Modern GPT adds RoPE and SwiGLU and reaches 94.4~$\pm$~0.2\%. The WS model reuses that Modern block for 8 passes and reaches 98.0\% validity, adding 3.6 points while using half the parameters of the unshared baselines (Appendix~\ref{app:configs}, Table~\ref{tab:arch-ablation}), consistent with the Pareto dominance of looped architectures at larger scale \citep{reid2021subformer, prairie2026parcae, zeitoun2026hyperloop}.

To test whether iteration explains the gain, we compare against three alternatives. First, removing iteration from WS-206K collapses validity by 65 points. Second, larger unshared GPTs remain below WS-206K. GPT-241K reaches 88.4~$\pm$~0.6\%, and GPT-228K-mod reaches 91.2~$\pm$~1.4\% (3 seeds each), despite having more parameters. Third, extra training compute does not close the gap. A matched Modern GPT reaches 94.2~$\pm$~0.1\% after 800 epochs (8 times longer, with roughly twice WS-206K's training compute), similar to its 100-epoch value and still 3.8 points below WS-206K. Thus, the compact model succeeds by refining SMILES validity over repeated applications of the same small block.

\subsection{Robustness checks}
\label{sec:robustness}

We test whether this compact-generation result survives changes in model class, SMILES augmentation, sampling settings, and post-training.

First, we compare against recurrent models at matched parameter count. GRU controls trail the weight-shared block, with GRU-53K reaching 87.7\% and GRU-206K reaching 95.5\% (2 seeds each, 25--30 training epochs vs.\ 100 for the WS comparators; matched-epoch comparison left to future work). At matched parameter count, WS exceeds the GRU by 7.7 points at 53K and 2.5 points at 206K, narrowing with capacity in line with larger GRUs ($>$1M) reaching high validity \citep{segler2018generating}. Thus, the GRU comparison supports the compact-generation result at matched parameter count, while showing that the advantage narrows as recurrent capacity increases.

Second, we reduce SMILES enumeration by training with one string representation per molecule instead of 10 randomized SMILES. Under this setting, WS-53K drops from 95.3\% to 92.3\%, a 3.0-point decrease, and WS-206K drops from 98.0\% to 97.8\%, a 0.2-point decrease within seed variance (3 seeds), so augmentation helps but less than architecture.

Third, we vary decoding temperature, top-$k$, and top-$p$. Across temperature ($T \in \{0.7, 0.8, 1.0, 1.2\}$), top-$k$ ($\{40, \infty\}$), and top-$p$ ($\{0.95, \text{none}\}$), WS-206K remains above the 403K unshared Modern GPT in every effective cell despite using half as many parameters (sign test $p{<}10^{-3}$ over 12 effective cells, since at $T{\le}0.8$ the top-$k{=}40$ and top-$k{=}\infty$ cells are bitwise-identical).

Finally, we test standard post-training methods. Offline knowledge distillation (KD; \citealp{hinton2015distilling}) drops WS-206K validity by ${\sim}8$ percentage points (3 seeds); the KL on soft labels spreads probability mass onto invalid tokens (Appendix~\ref{app:kd_ablation}), and teacher quality has no effect (Appendix~\ref{app:teacher}). Direct Preference Optimization \citep[DPO;][]{rafailov2024direct} preserves validity to within 1~pp (WS-206K $98.5{\rightarrow}97.6\%$; full results in Appendix~\ref{app:dpo}), so the compact-generation result is robust to preference fine-tuning. Together, these robustness checks support the interpretation that compact validity comes from repeated computation through the shared block, not from the sampler, memorization, or post-training.

\section{Computation across passes}
\label{sec:mechanism}

We next ask \textit{how} weight-shared models generate valid molecules across repeated passes. Reusing a block provides iterative computation, but it does not determine whether successive passes perform distinct computations. We find here a fixed hierarchy of increasing scope. The shared block resolves brackets by pass 2 (local pairings within a few tokens), rings by pass 4 (long-range digit pairing across tens of tokens), and valence by pass 8 (atom-level integration of full bond context).

The rest of this section establishes this hierarchy with four complementary measurements: outputs, representations, causal structure, and capacity. We first classify errors across virtual depth to test whether failures disappear in the bracket-to-ring-to-valence order (Section~\ref{sec:error}). We then use linear probes and sparse autoencoders to ask whether hidden states follow the same progression from token-level features to bracket depth, ring state, and atom identity (Section~\ref{sec:representation}). We next ablate individual heads to localize bracket matching to a single attention head in WS-206K (Section~\ref{sec:ablation}). Finally, we compare model sizes to test whether smaller blocks resolve the same constraints slightly later and with less localized circuitry while preserving the order itself (Section~\ref{sec:capacity}).

\subsection{Errors resolve in order}
\label{sec:error}

To test whether repeated passes specialize at the output level, we classify errors in invalid molecules generated at each virtual depth. This sweep isolates virtual depth by holding block size fixed and varying $D$ in separately trained WS-206K-D$N$ models. Later analyses hold $D{=}8$ and compare WS-53K with WS-206K. Specialization should not only make invalid samples rarer, but it should also shift failure modes from simpler syntax errors toward harder chemical constraints. The three error classes instantiate this order: bracket errors test parenthesis matching, ring errors test long-range digit pairing, and valence errors test context-sensitive chemical consistency. We train a separate model at each depth ($D \in \{1,2,4,8,16\}$), which avoids reading early-pass representations through an output head calibrated for pass 8.

\begin{table}[t]
\centering
\caption{Error classification for separately-trained WS-206K-D$N$ models at depths $N \in \{1,2,4,8,16\}$, with 1{,}000 samples per depth over 3 seeds ($\pm$ is standard deviation). Each invalid molecule is assigned to one failure class (Brackets, Rings, Valence, Other), rather than counting all failures in the string.}
\label{tab:errors}
\small
\begin{tabular}{lccccc}
\toprule
Passes & Valid (\%) & Brackets (\%) & Rings (\%) & Valence (\%) & Other (\%) \\
\midrule
1  & 33.0 $\pm$ 2.3 & 23.0 $\pm$ 1.7 & 12.8 $\pm$ 1.4 & 31.0 $\pm$ 1.3 & 0.3 $\pm$ 0.2 \\
2  & 90.5 $\pm$ 1.8 &  1.3 $\pm$ 0.5 &  1.9 $\pm$ 0.3 &  6.2 $\pm$ 1.2 & 0.0 $\pm$ 0.1 \\
4  & 97.1 $\pm$ 0.6 &  0.1 $\pm$ 0.1 &  0.8 $\pm$ 0.3 &  2.0 $\pm$ 0.9 & 0.1 $\pm$ 0.1 \\
8  & 98.6 $\pm$ 0.2 &  0.1 $\pm$ 0.1 &  0.3 $\pm$ 0.2 &  1.0 $\pm$ 0.3 & 0.0 $\pm$ 0.1 \\
16 & 98.7 $\pm$ 0.3 &  0.0 $\pm$ 0.1 &  0.5 $\pm$ 0.1 &  0.7 $\pm$ 0.3 & 0.1 $\pm$ 0.1 \\
\bottomrule
\end{tabular}
\end{table}

Table~\ref{tab:errors} shows the predicted shift in assigned failure modes. At 1 pass, many samples fail before reaching later checks. By pass 2, bracket errors have collapsed from 23.0\% to 1.3\%, consistent with early resolution of parenthesis matching. Among the remaining assigned failures, ring-closure errors then drop between passes 2 and 4 (1.9\% to 0.8\%), resolving long-range digit pairing. Valence violations are the last substantial residual class, dropping from 2.0\% to 1.0\% between passes 4 and 8. The main error sweep therefore supports the predicted bracket-to-ring-to-valence sequence.

The main sweep compares separately trained models, so the ordering could in principle reflect training a different model at each depth rather than computation inside one model. To control for this depth-training confound, we read the trained 8-pass model's output head at each pass (Table~\ref{tab:errors-readout}). This intermediate readout gives lower absolute validity, as expected when a pass-8 head is applied to earlier representations, but the ordering is unchanged, with brackets the first error type to drop below 5\% (pass 3), rings the second (pass 5), and valence the third (pass 6). Thus, the intermediate readout rules out training depth as the source of the ordering.

\subsection{Representations organize across passes}
\label{sec:representation}

The output-level analysis shows when each constraint is satisfied in the generated molecules, and we test whether the same hierarchy appears inside the model's hidden states as the shared block is applied repeatedly. We use two complementary tests on the hidden states of the trained 8-pass model (pass $k$ refers to the state after the $k$-th application of the block). First, linear probes ask whether bracket depth and ring state are decodable at each pass. Second, sparse autoencoders (SAEs) ask whether interpretable chemical features emerge without specifying a readout in advance. Table~\ref{tab:representation} summarizes both analyses; they show the same bracket-then-ring ordering.

\begin{table}[ht]
\centering
\caption{Representation organization across passes. Probing rows report peak linear-probe accuracy. SAE rows report the peak feature--property correlation $r$ across all learned features. Per-pass curves appear in Fig.~\ref{fig:representation}.}
\label{tab:representation}
\small
\begin{tabular}{llcccc}
\toprule
& & \multicolumn{2}{c}{WS-53K} & \multicolumn{2}{c}{WS-206K} \\
\cmidrule(lr){3-4} \cmidrule(lr){5-6}
Method & Property & peak & pass & peak & pass \\
\midrule
\multirow{2}{*}{Probing}
 & Bracket depth        & 98.4\% & 3 & 99.2\% & 3 \\
 & Ring state           & 98.0\% & 5 & 98.6\% & 4 \\
\midrule
\multirow{5}{*}{SAE}
 & Bracket detector     & 0.97 & 2 & 0.80 & 1 \\
 & Ring-digit detector  & 0.99 & 3 & 0.93 & 3 \\
 & Bracket depth        & 0.64 & 2,\,8 & 0.55 & 4 \\
 & Atom identity        & 0.84 & 2 & 0.62 & 1 \\
 & Ring state           & 0.64 & 5 & 0.75 & 4 \\
\bottomrule
\end{tabular}
\end{table}

\paragraph{Linear probing.}

We train linear probes on each pass of the 8-pass model to test when bracket depth and ring state become decodable. In both models, probes decode bracket depth earlier than ring state. WS-53K reaches 98.4\% bracket-depth accuracy at pass 3, before ring-state accuracy reaches 98.0\% at pass 5. WS-206K reaches 99.2\% bracket-depth accuracy at pass 3, before ring-state accuracy reaches 98.6\% at pass 4. Although WS-53K starts 3--6 points lower than WS-206K at pass 1, both models reach comparable probe accuracy by pass 5, so the difference is timing rather than final decodability. The ordering is the same in a single trained model as in the output-level analysis.

\paragraph{Sparse autoencoders.}

Sparse autoencoders test the same progression without training a linear probe for each property. For each pass, an SAE learns sparse features from hidden states. We then ask which features track five chemical properties: \texttt{is\_bracket}, \texttt{is\_ring\_digit}, \texttt{bracket\_depth}, \texttt{ring\_state}, and \texttt{is\_atom} (Appendix~\ref{app:representation}). The learned features separate by pass most clearly in WS-53K. Token-level detectors (\texttt{is\_bracket}, \texttt{is\_ring\_digit}) appear early, at passes 2--3 with correlations $r{=}0.97$ and $0.99$. Atom identity also achieves high correlation early (pass 2 in WS-53K, $r{=}0.84$), consistent with atom-type information being available from early passes. Two features require integrating information across multiple tokens: bracket \emph{depth} (the count of unmatched open brackets at each position) and ring \emph{state} (which ring digits are currently open). Ring-state correlation reaches its peak at pass 5 ($r{=}0.64$). Bracket-depth correlation is bimodal: near-equal peaks at passes 2 and 8 ($r{=}0.63$ and $0.64$), separated by a dip to $r{\approx}0.49$. WS-206K shows weaker separation (Table~\ref{tab:representation}; Fig.~\ref{fig:representation}). Because the SAEs are not trained to predict these properties, the features appear as emergent decompositions rather than only as supervised readouts.

Moreover, feature--property correlations are stable across corpus size (Appendix~\ref{app:representation}). When we scale the SAE corpus from 2{,}000 to 20{,}000 ZINC-250K SMILES, WS-53K feature--property correlations strengthen or remain within 0.01. Taken together, error classification, supervised probes, and the SAE support the same brackets-to-rings-to-valence hierarchy.

\subsection{The bracket-matching head}
\label{sec:ablation}

In both error classification and internal representations, bracket matching is the first constraint to resolve. We further test whether that first step can be traced to a specific component of the shared block. If so, ablating one attention head at one pass should selectively raise bracket errors, rather than simply lowering validity whenever any part of the block is removed. We ablate a head by zeroing its contribution at one pass, then measuring the full error profile. We repeat this intervention across all four heads and all eight passes, for a total of 32 head-pass combinations, and generate $n{=}2{,}000$ molecules per condition for both models. Table~\ref{tab:ablation} summarizes the resulting changes in bracket errors and validity.

\begin{table}[t]
\centering
\caption{Head-ablation sweep for WS-53K and WS-206K. Changes are percentage points (pp) from the unablated baseline ($n{=}2{,}000$ per condition). The bracket-matching head is identified per seed (WS-206K heads 1/2/0, WS-53K heads 3/1/3 across seeds 42/43/44) and shown separately. Other heads and passes are aggregated.}
\label{tab:ablation}
\small
\begin{tabular}{lcccc}
\toprule
& \multicolumn{2}{c}{WS-53K} & \multicolumn{2}{c}{WS-206K} \\
\cmidrule(lr){2-3} \cmidrule(lr){4-5}
Ablation & $\Delta$brackets (pp) & $\Delta$validity (pp) & $\Delta$brackets (pp) & $\Delta$validity (pp) \\
\midrule
Bracket head, pass 1         & $+$10 & $-$10 & $+$23 & $-$23 \\
Bracket head, pass 2         & $+$5  & $-$6  & $+$17 & $-$17 \\
Bracket head, pass 3         & $+$2  & $-$2  & $+$5  & $-$5 \\
Bracket head, pass 4+        & $<$2  & $<$2  & $<$2  & $<$2 \\
All other heads, all passes  & $<$2  & $<$2  & $<$2  & $<$2 \\
\bottomrule
\end{tabular}
\end{table}

Across the full sweep, we localize the bracket-matching computation to a single attention head in WS-206K. Ablating that head at pass 1 raises bracket errors by 23 points while ring and valence errors remain at baseline; the effect persists through passes 2--3 (increases of 17 and 5 points) and then vanishes by pass 4, and all other heads at every pass remain near baseline. The dependence is thus specific to one head rather than to early-pass computation in general. WS-53K does not concentrate bracket matching the same way: at seed 42, no single head dominates (the largest pass-1 bracket-error increase is 3 points), and the 10-point increase in the 3-seed mean is carried by different heads in different seeds (heads 3, 1, and 3 for seeds 42, 43, and 44). The smaller model therefore distributes bracket matching across heads, whereas the larger model localizes it to one.

We use two controls to test whether the effect is specific to one head rather than to early-pass ablation or activation magnitude in general. To rule out general early-pass sensitivity, we ablate all of pass 1 in WS-206K rather than the bracket head alone; this leaves bracket errors at baseline and drops validity by only 0.5 points, roughly 50 times smaller than the 23-point drop from the single-head ablation. To rule out magnitude-driven effects, we compare activation norms: in all three seeds, the bracket head has the \emph{smallest} pass-1 norm, and ablating the \emph{largest}-norm head does not produce a larger validity drop. Together, the controls show that bracket matching depends on this specific head.

\subsection{Smaller models need more passes}
\label{sec:capacity}

To test whether the pass-by-pass hierarchy changes with capacity, we compare WS-53K and WS-206K across error classification and probing. We find that WS-53K follows the same sequence as WS-206K but reaches its resolution points about one pass later (Table~\ref{tab:capacity}): per-pass validity peaks at pass 8 rather than pass 7, and the ring-state probe peaks at pass 5 rather than pass 4. Bracket depth peaks at pass 3 in both models, so the shift is small and appears only in the longer-range constraints. The two models reach comparable final accuracies (95.3\% vs.\ 98.0\% validity, and 98.4\% vs.\ 99.2\% best bracket-probe accuracy). Capacity also changes how bracket matching is implemented: WS-206K localizes it to a single head (Section~\ref{sec:ablation}), whereas WS-53K distributes it across heads. Thus, lower capacity slightly delays when each constraint resolves and spreads bracket matching across more heads, but does not change the order in which the constraints resolve.

\section{Discussion}

This paper shows that weight sharing can make molecular generation both compact and interpretable. With only 53K parameters, a shared-block transformer generates highly valid SMILES while preserving a visible sequence of validity computation. This improves the parameter-validity tradeoff, showing that small models can recover much of the validity of substantially larger unshared transformers.

We observe the same sequence across generated errors, hidden-state probes, and ablations: bracket matching first, ring closure second, and valence last. This consistency suggests that the shared block does more than reduce parameter count. By reusing one set of weights across passes, the model exposes successive stages of validity computation.

The ordering also links the empirical result to the grammar of SMILES. Bracket matching, ring closure, and valence are not arbitrary metrics; they form a progression from local syntax to long-range structure to chemically constrained validity. This connects the observed sequence to formal-language results in which fixed-depth transformers struggle with context-free and higher languages, while iterative depth supplies the recurrence such constraints require \citep{hahn2020theoretical,deletang2023neural}. Weight sharing makes this connection visible because the same parameters are applied repeatedly, so cross-pass changes track computation through time rather than differences between learned layers.

\paragraph{Limitations.}

We qualify the claims in three ways. First, the empirical scope is deliberately focused. We study ZINC-250K SMILES, so the ordering should be read as a claim about validity computation in this representation rather than about molecular structure independent of representation. Second, the mechanistic measurements are finite-sample estimates. The SAE analysis uses 20{,}000 SMILES (${\sim}$770K tokens), so rare features may only emerge at larger scale, and the ablation sweep uses 2{,}000 molecules per condition, enough to identify the bracket-head effect but not to resolve very small effects from other heads. Third, weight sharing reduces parameters but not autoregressive inference cost, because FLOPs scale with pass count.

\paragraph{Broader impacts.}

Compact molecular generators can lower the compute cost of exploratory molecule design and mechanism analysis. The same capability could be misused to propose molecules with undesirable biological activity, so generated molecules should not be used for synthesis prioritization without toxicity, safety, and feasibility screening.

\paragraph{Future directions.}

These results point to broader tests of whether shared depth can expose structured computation in other scientific domains. One direction is to extend to other biological, chemical, and materials systems where similar hierarchies of constraints arise, such as RNA folding from local base pairing to long-range tertiary contacts. A second direction is to span spatial and temporal scales by letting each pass resolve a different scale, from atomic interactions to macroscale dynamics and mechanics, providing a single architecture for problems that today require stitching separate models together. A third is to design distillation objectives that preserve the iterative computation, for example by matching per-pass representations rather than only output distributions.

\section{Related work}

\paragraph{Molecular generation with language models.}
SMILES generation has progressed from RNN-based models \citep{segler2018generating, olivecrona2017molecular} to transformer decoders such as MolGPT \citep{bagal2022molgpt}, GP-MoLFormer \citep{ross2024gpmolformer}, and REINVENT~4 \citep{loeffler2024reinvent}. Recent scaling work studies five molecular representations from 1M to 650M parameters \citep{xu2026scaling}. Its smallest model is 19 times larger than WS-53K. To our knowledge, no prior transformer work explores molecular generation below 1M parameters, the regime where reusing computation offers a direct alternative to scaling parameter count.

\paragraph{Weight sharing and looped transformers.}
Prior work uses weight sharing to compress transformers and to turn depth into repeated computation. Compression models such as ALBERT \citep{lan2020albert}, Subformer \citep{reid2021subformer}, MobileLLM \citep{liu2024mobilellm}, and Relaxed Recursive Transformers \citep{bae2025relaxed} apply shared blocks at scales 1,000 to 10,000 times larger, where shared models match unshared baselines with fewer parameters \citep{prairie2026parcae, zeitoun2026hyperloop}. DeLiCaTe \citep{yu2022delicate} brings the same compression motive to SMILES property prediction. Other work uses repeated depth as computation, including the Universal Transformer \citep{dehghani2019universal}, theoretical results on looped transformers \citep{giannou2023looped}, and recent looped models with iteration-wise specialization \citep{yu2026spiralformer,goyal2026elt}. We use the same architectural device for a different purpose. Because SMILES validity depends on repeated grammar checks, weight sharing makes those checks observable as computation through time.

Large looped language models make repeated computation central, but they leave individual iterations difficult to interpret. Models such as Huginn \citep{geiping2025huginn,lu2025huginn} and Ouro \citep{zhu2025ouro} operate at billion-parameter scale, where different iterations can perform similar computations \citep{blayney2026mechanistic}. Natural language also lacks the exact symbolic validity checks available in SMILES. Our setting combines a sub-100K bottleneck with constraints that can be measured at every pass, so the iteration is easier to separate into steps. This complements work on thermodynamic phases in looped transformers \citep{lam2026energy}, formal-language constraints and circuit complexity \citep{hu2025circuits}, and concurrent studies of per-step specialization in latent reasoning models \citep{li2026latent,kohli2026looping,kaissis2026perstep}. Where those studies rely mainly on correlational traces, training stages, or trace decoding \citep{dilgren2026latent}, SMILES lets us test each intermediate computation with symbolic checks and causal ablation.

\paragraph{Mechanistic interpretability.}
Mechanistic interpretability provides tools for finding circuits and features, but standard layerwise analyses make temporal ordering hard to separate from layer identity. We combine sparse autoencoders \citep{bricken2023monosemanticity, cunningham2024sparse}, causal ablation \citep{vig2020causal}, and probing \citep{conneau2018probing}. \citet{varadi2025circuits} identify branch-balancing and ring-closure circuits in a 25M-parameter molecular transformer, establishing that chemically meaningful SMILES circuits can be recovered with SAE-based analysis. Weight sharing lets us test how related constraints resolve across repeated applications of the same block, rather than only where they appear across layers with separate weights. This distinction matters because \citet{tenney2019bert} find a correlational syntax-to-semantics arc across BERT layers, while subsequent work questions such arcs under controls for layer position \citep{niu2022does}. Shared weights remove that layer-identity confound. We also use causal ablation because attention-based analysis of bracket matching can mislead \citep{wen2023dyck}, and we validate SAE features against symbolic chemical ground truth rather than auto-interpretability scores \citep{heap2025saerandom}. This per-pass analysis complements SMI-TED's single-extraction-point SAE \citep{cohen2025smited} and TIDE's temporal SAEs for diffusion transformers \citep{huang2026tide}.

\section{Conclusion}

SMolLM, a 53K-parameter weight-shared transformer, reaches 95.3\% validity on ZINC-250K, outperforms an unshared GPT with 10 times more parameters, and shifts the sub-megaparameter Pareto frontier. The architecture is also mechanistically interpretable. Across generated errors, hidden-state probes, and ablations, the shared block resolves SMILES constraints in a fixed pass-by-pass hierarchy (brackets by pass 2, rings by pass 4, valence by pass 8), consistent from 53K to 206K parameters, with capacity changing how quickly each constraint resolves rather than the order itself. Taken together, weight sharing is not only a compression device for molecular generation but also a way to study iterative computation in domains with exact symbolic constraints.

\bibliographystyle{plainnat}
\bibliography{ref}

\newpage
\appendix
\section*{Appendix}
\addcontentsline{toc}{section}{Appendix}

\section{Architecture configurations}
\label{app:configs}

\begin{table}[h]
\centering
\caption{Architecture configurations, parameter counts, and reproducibility identifiers. \emph{Name} is the canonical identifier used throughout the paper; \emph{checkpoint ID} is the matching filename under \texttt{checkpoints/0504\_paper/} on Hugging Face (\url{https://huggingface.co/akhljndl/smollm}) and the corresponding wandb run name (entity \texttt{ajindal}, project \texttt{smollm}, tag \texttt{0504\_paper}). Class is the block family (Standard GPT, Modern GPT with RoPE+SwiGLU, weight-shared Modern GPT, or GRU). The \texttt{-mod} suffix on a GPT name marks the Modern block; the \texttt{-D$N$} suffix on a WS name marks a non-default virtual depth (default $D{=}8$); the \texttt{-8ep} suffix marks the 800-epoch training-compute control (default 100 epochs).}
\label{tab:configs}
\small
\setlength{\tabcolsep}{4pt}
\begin{tabular}{lllccccc}
\toprule
Name & Class & Checkpoint ID & $L$ & $H$ & $E$ & $D$ & Params \\
\midrule
GPT-111K       & Standard GPT  & \texttt{L2H2E64}        & 2 & 2 & 64  & --  & 111K \\
GPT-419K       & Standard GPT  & \texttt{L2H2E128}       & 2 & 2 & 128 & --  & 419K \\
GPT-527K       & Standard GPT  & \texttt{L2H2E144}       & 2 & 2 & 144 & --  & 527K \\
GPT-813K       & Standard GPT  & \texttt{L4H4E128}       & 4 & 4 & 128 & --  & 813K \\
GPT-3.2M       & Standard GPT  & \texttt{gpt-3.2m}       & 4 & 4 & 256 & --  & 3.2M \\
GPT-10.7M      & Standard GPT  & \texttt{L6H6E384}       & 6 & 6 & 384 & --  & 10.7M \\
GPT-419K-h4    & Standard GPT  & \texttt{L2-E128-vanilla}& 2 & 4 & 128 & --  & 419K \\
GPT-241K       & Standard GPT  & \texttt{L2-E96-vanilla} & 2 & 4 & 96  & --  & 241K \\
GPT-403K-mod   & Modern GPT    & \texttt{L2-E128-modern} & 2 & 4 & 128 & --  & 403K \\
GPT-403K-mod-8ep & Modern GPT  & \texttt{L2-E128-modern-8ep} & 2 & 4 & 128 & --  & 403K \\
GPT-228K-mod   & Modern GPT    & \texttt{L2-E96-modern}  & 2 & 4 & 96  & --  & 228K \\
GPT-467K-mod   & Modern GPT    & \texttt{L6-E80-modern}  & 6 & 4 & 80  & --  & 467K \\
WS-15K         & Weight-shared & \texttt{ws-e32-x8}      & 1 & 4 & 32  & 8   & 15K  \\
WS-31K         & Weight-shared & \texttt{ws-e48-x8}      & 1 & 4 & 48  & 8   & 31K  \\
WS-53K         & Weight-shared & \texttt{ws-e64-x8}      & 1 & 4 & 64  & 8   & 53K  \\
WS-82K         & Weight-shared & \texttt{ws-e80-x8}      & 1 & 4 & 80  & 8   & 82K  \\
WS-115K        & Weight-shared & \texttt{ws-e96-x8}      & 1 & 4 & 96  & 8   & 115K \\
WS-156K        & Weight-shared & \texttt{ws-e112-x8}     & 1 & 4 & 112 & 8   & 156K \\
WS-206K        & Weight-shared & \texttt{ws-e128-x8}     & 1 & 4 & 128 & 8   & 206K \\
WS-206K-D$N$   & Weight-shared & \texttt{ws-e128-x$N$}   & 1 & 4 & 128 & $N{\in}\{1,2,4,16\}$ & 206K \\
GRU-53K        & GRU           & \texttt{gru-53k}        & -- & -- & 64  & --  & 54K  \\
GRU-206K       & GRU           & \texttt{gru-206k}       & -- & -- & 105 & --  & 207K \\
\bottomrule
\end{tabular}
\end{table}

\begin{table}[H]
\centering
\caption{Validity and FCD across the size sweep (companion to Fig.~\ref{fig:scaling}). Validity cells with $\pm$ are means over 3 seeds; cells without are single-seed.}
\label{tab:pareto}
\small
\setlength{\tabcolsep}{6pt}
\renewcommand{\arraystretch}{1.05}
\begin{tabular}{lrcc}
\toprule
Model & Params & Validity (\%) & FCD \\
\midrule
\multicolumn{4}{l}{\emph{Standard GPT}} \\
\quad GPT-111K   & 111K & 62.1 & 4.58 \\
\quad GPT-419K   & 419K & 84.8 $\pm$ 0.7 & 3.03 \\
\quad GPT-527K   & 527K & 87.6 $\pm$ 0.7 & 3.01 \\
\quad GPT-813K   & 813K & 98.7 $\pm$ 0.2 & 2.44 \\
\quad GPT-3.2M   & 3.2M & 99.4 $\pm$ 0.1 & 2.32 \\
\quad GPT-10.7M  & 10.7M & 99.7 & 2.08 \\
\midrule
\multicolumn{4}{l}{\emph{Weight-shared (ours)}} \\
\quad WS-15K   & 15K  & 81.2 $\pm$ 1.0 & 4.57 \\
\quad WS-31K   & 31K  & 91.7 $\pm$ 0.5 & 3.10 \\
\quad WS-53K   & 53K  & 95.3 $\pm$ 0.7 & 2.76 \\
\quad WS-82K   & 82K  & 96.2 $\pm$ 0.4 & 2.66 \\
\quad WS-115K  & 115K & 97.3 $\pm$ 0.3 & 2.57 \\
\quad WS-156K  & 156K & 98.3 $\pm$ 0.2 & 2.51 \\
\quad WS-206K  & 206K & 98.0 $\pm$ 0.4 & 2.45 \\
\bottomrule
\end{tabular}
\end{table}

\begin{table}[H]
\centering
\caption{Architecture ablation (block type varies; Standard~$=$ learned positional embeddings $+$ GELU; Modern~$=$ RoPE $+$ SwiGLU). Each row adds one architectural modification; the final row removes iteration to isolate its contribution. WS halves parameters yet delivers the largest jump. Baselines use $H{=}4$ heads to match the weight-shared block, isolating iteration from head count; this differs from Table~\ref{tab:pareto}'s GPT-419K ($H{=}2$, same embedding dimension and parameter count).}
\label{tab:arch-ablation}
\small
\setlength{\tabcolsep}{6pt}
\renewcommand{\arraystretch}{1.05}
\begin{tabular}{lrcc}
\toprule
Model & Params & Validity (\%) & $\Delta$ (pp) \\
\midrule
GPT-419K-h4 (Standard)   & 419K & 91.6 $\pm$ 0.3 & -- \\
GPT-403K-mod (Modern)    & 403K & 94.4 $\pm$ 0.3 & $+$2.8 \\
\quad + weight sharing (WS-206K)  & 206K & 98.0 $\pm$ 0.4 & $+$3.6 \\
\midrule
\multicolumn{4}{l}{\emph{Capacity-matched unshared baselines (\,$\geq$ WS-206K's 206K)}} \\
GPT-241K (Standard)      & 241K & 88.4 $\pm$ 0.6 & -- \\
GPT-228K-mod (Modern)    & 228K & 91.2 $\pm$ 1.4 & -- \\
\midrule
\emph{Iteration removed} (WS-206K-D1) & 206K & 33.0 $\pm$ 2.3 & $-$65.0 (vs.\ WS-206K) \\
\bottomrule
\end{tabular}
\end{table}

\section{Training and evaluation details}
\label{app:setup}

\textbf{Dataset.} ZINC-250K \citep{irwin2012zinc}: 249{,}455 drug-like molecules. SMILES augmentation \citep{bjerrum2017smiles}: each canonical SMILES expanded to $n_\text{aug}=10$ randomized SMILES via RDKit. Character-level tokenizer, vocabulary size 50, max sequence length 128.

\textbf{Training.} 100 epochs, AdamW (weight decay 0.1), batch size 256, cosine LR (peak $5 \times 10^{-4}$, 200-step warmup), mixed-precision training (AMP), gradient clipping 1.0, no label smoothing. Distillation (Section~\ref{sec:robustness}): offline KD with softmax temperature $\tau=2.0$ and hard/soft loss mixing coefficient $\alpha=0.5$; teacher GPT-467K-mod (98.5\%). DPO (Section~\ref{sec:robustness}): 5 epochs, lr $=10^{-5}$, KL regularization strength $\beta=0.1$, sample budget 10K molecules per base, pairing valid samples against invalid ones (so usable pairs are bounded by the invalid count). Full DPO results are in Appendix~\ref{app:dpo}.

\textbf{Evaluation.} $N=1{,}000$ molecules per model, temperature $T=0.8$, top-$k=40$. Metrics: validity (RDKit parsing), FCD (Fr\'echet ChemNet Distance, analogous to FID; \citealp{preuer2018fcd}), internal diversity ($1 - $ mean pairwise Tanimoto, Morgan fingerprints radius 2, 1024 bits), uniqueness, novelty.

\textbf{Compute.} Reported runs used single AMD Instinct MI210 or MI250 GPUs (64--128GB HBM class) on an internal cluster. Training time ranged from under 2 GPU-hours for the smallest 53K models to under 24 GPU-hours for the largest reported models; most sub-megaparameter runs completed in a few GPU-hours. The reported experiments used approximately 500 GPU-hours total, excluding preliminary and failed runs that were not used in the paper. Checkpoints and logs require less than 1GB per run.

\textbf{Pass indexing.} The paper uses 1-indexed pass numbers (pass 1 is the first block application; pass 8 is the last). The released code uses 0-indexed positions (\texttt{pass=0..7}); to map paper to code, subtract 1.

\textbf{Reproducibility.} Code is released at \url{https://github.com/akhljndl/smollm} (MIT). Each experiment maps to a numbered configuration in \texttt{experiments/smiles\_gen/configs.py} and can be reproduced via \texttt{python -m experiments.smiles\_gen.run --exp <N>}. Trained checkpoints (3 seeds per config, all 21 architectures) are on Hugging Face at \url{https://huggingface.co/akhljndl/smollm} under \texttt{checkpoints/0504\_paper/}, and training runs are public on Weights \& Biases (entity \texttt{ajindal}, project \texttt{smollm}, tag \texttt{0504\_paper}). Two stability notes: WS-206K's 3-seed std is 0.33 percentage points, and run-to-run variance across independent training runs is consistent with this. The four Table~\ref{tab:arch-ablation} checkpoints were regenerated and verified within 2 percentage points of reported values at $N{=}10{,}000$ samples.

\section{Error classification across passes}
\label{app:errors}

Companion plot to Table~\ref{tab:errors}: error-type composition across virtual depths in separately-trained WS-206K-D$N$ models. With $n{=}1{,}000$ samples per depth averaged over 3 seeds, the cross-pass trends are robust.

\begin{figure}[!htbp]
\centering
\includegraphics[width=0.78\textwidth]{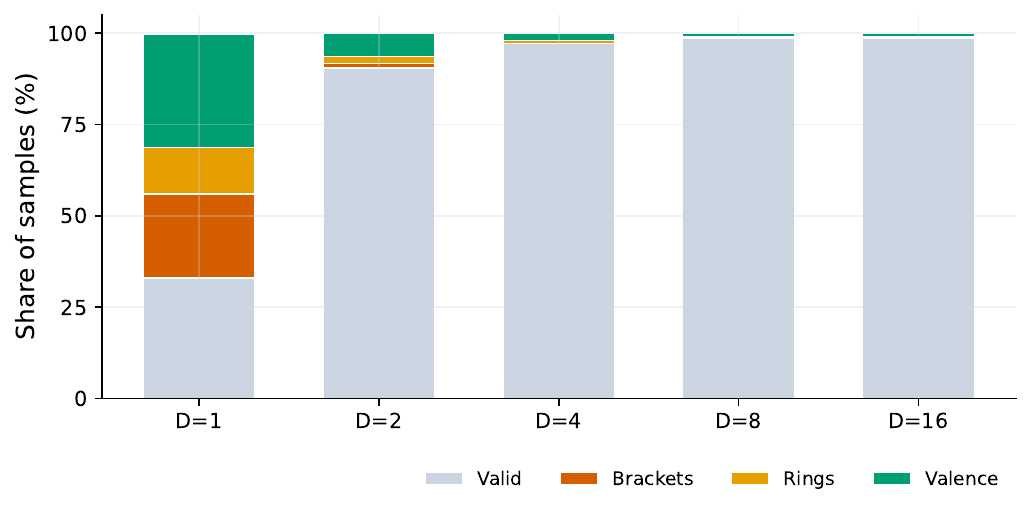}
\caption{Bracket errors collapse by pass 2, rings by pass 4, valence by pass 8 (companion to Table~\ref{tab:errors}).}
\label{fig:errors}
\end{figure}

\begin{table}[H]
\centering
\caption{Per-pass readout from the trained 8-pass WS-206K model, averaged over 3 seeds ($n{=}500$ per pass per seed). Same constraint ordering as Table~\ref{tab:errors} (brackets$\rightarrow$rings$\rightarrow$valence); absolute validities are head-calibration-shifted. Valid reports the fraction of generated molecules that pass all checks. Brackets, Rings, Valence, and Other assign each invalid molecule to one failure class, rather than counting all possible failures in the same string.}
\label{tab:errors-readout}
\small
\begin{tabular}{lccccc}
\toprule
Pass & Valid (\%) & Brackets (\%) & Rings (\%) & Valence (\%) & Other (\%) \\
\midrule
1 &  6.7 & 36.1 & 45.1 & 11.6 & 0.5 \\
2 & 20.1 & 12.1 & 45.1 & 22.3 & 0.4 \\
3 & 51.0 &  3.1 & 22.1 & 23.3 & 0.4 \\
4 & 76.4 &  0.8 &  8.8 & 13.8 & 0.2 \\
5 & 91.5 &  0.2 &  2.2 &  6.0 & 0.1 \\
6 & 96.8 &  0.0 &  0.7 &  2.5 & 0.0 \\
7 & 98.7 &  0.0 &  0.3 &  0.9 & 0.1 \\
8 & 98.4 &  0.0 &  0.5 &  1.1 & 0.1 \\
\bottomrule
\end{tabular}
\end{table}

\FloatBarrier
\section{Head-ablation heatmap}
\label{app:ablation}

Companion to Table~\ref{tab:ablation}: full head $\times$ pass heatmap for both models. We apply a Bonferroni correction across the $K{=}32$ head-pass ablations for each model.

\begin{figure}[H]
\centering
\includegraphics[width=\textwidth]{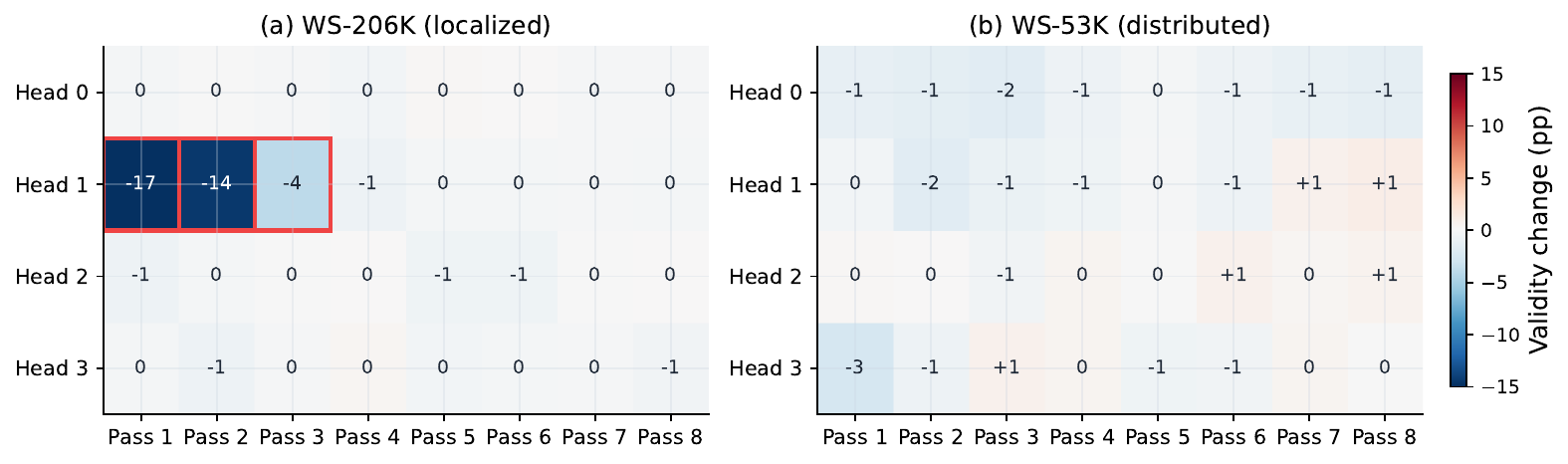}
\caption{Ablation heatmap (head $\times$ pass, change in validity in percentage points; seed 42, $n{=}2{,}000$ per condition). WS-206K concentrates the damage in one head (head 1) at passes 1--3 and is negligible at every other head and pass. WS-53K shows no comparable single-head effect at this seed, consistent with bracket matching being distributed across heads in the smaller model (3-seed-mean bracket-head effect $-$10 pp; Table~\ref{tab:ablation}).}
\label{fig:ablation}
\end{figure}

\FloatBarrier
\section{Representation organization across passes}
\label{app:representation}

Panel (a) shows linear-probe accuracy for bracket depth and ring state across passes. Panel (b) shows peak SAE feature correlation for five chemical properties. Companion to Table~\ref{tab:representation}.

For each pass, we train an SAE with 512 features (4 times the hidden dimension for WS-206K and 8 times for WS-53K) using a sparsity penalty on 20{,}000 ZINC-250K SMILES (${\sim}$770K tokens). Decoder dictionary atoms are constrained to unit $\ell_2$ norm following \citet{bricken2023monosemanticity}; the constraint reduces over the input dimension of the decoder weight so that each atom (column) has unit norm, which is necessary for cross-pass comparison of feature--property correlations under L1 sparsity. Scaling from 2{,}000 to 20{,}000 ZINC-250K SMILES increases or preserves the WS-53K feature--property correlations across all properties. The exact peak pass for an individual feature varies by $\pm 2$ passes across seeds. For the late-pass convergence analysis, we compute cosine similarity between consecutive hidden states from the same trained 8-pass WS-206K model; pass 7--8 similarity reaches 0.995 while generated validity remains 97--98\%.

\begin{figure}[!htbp]
\centering
\begin{subfigure}[b]{\textwidth}
\centering
\includegraphics[width=0.78\textwidth]{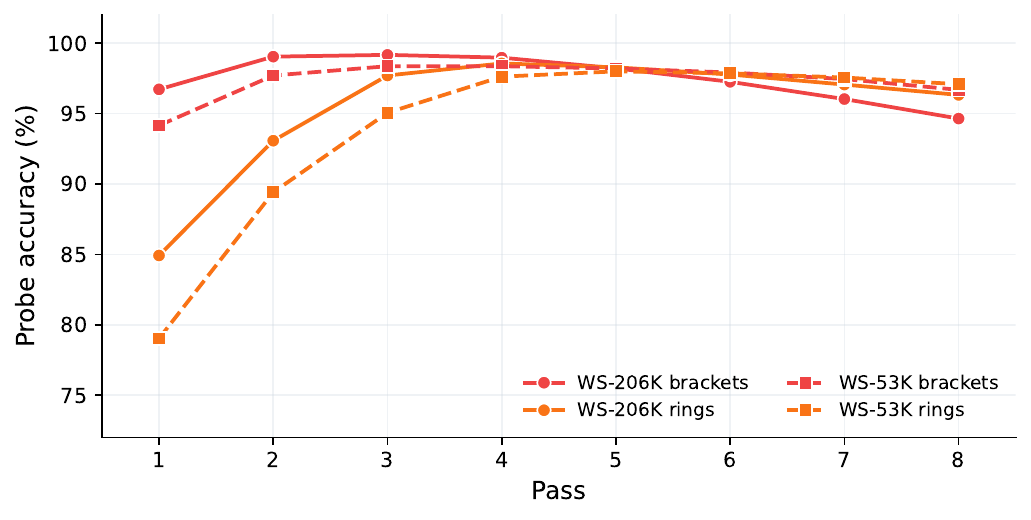}
\caption{Bracket depth saturates 1--2 passes before ring state in both models.}
\label{fig:probing}
\end{subfigure}

\vspace{0.6em}
\begin{subfigure}[b]{\textwidth}
\centering
\includegraphics[width=\textwidth]{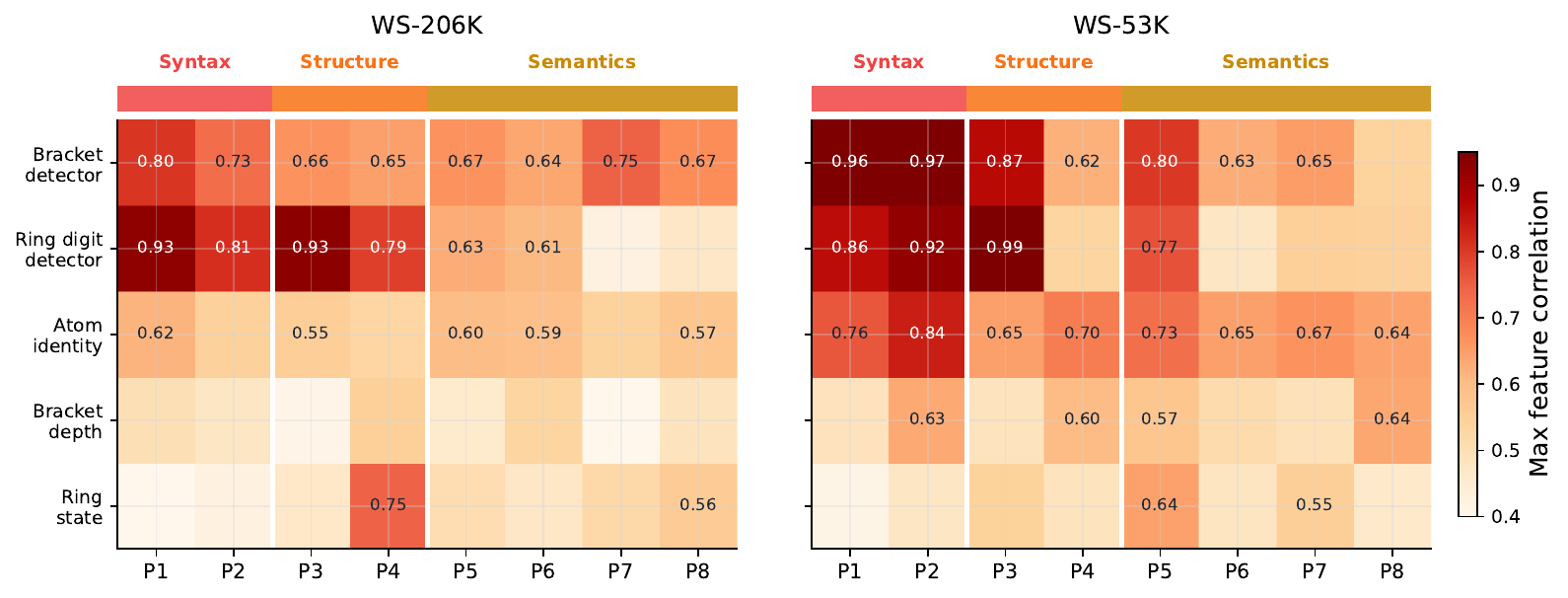}
\caption{SAE cross-pass correlation heatmap (properties $\times$ passes) for WS-53K and WS-206K. Blank cells indicate peak feature correlation $r < 0.4$. All five properties reach $r \geq 0.6$ in at least one pass for WS-53K, and four of five for WS-206K.}
\label{fig:sae}
\end{subfigure}
\caption{Representation organization across passes. Panel (a) shows probe accuracy; panel (b) shows SAE feature correlation. Companion to Table~\ref{tab:representation}.}
\label{fig:representation}
\end{figure}

\FloatBarrier
\section{Capacity comparison across methods}
\label{app:capacity}

The diagnostics agree that WS-53K solves the same subtasks as WS-206K, with peaks appearing at the same pass or slightly later (Table~\ref{tab:capacity}). Five of the six diagnostic rows shift by 0 to $+$1 pass, with no negative shifts; the exception is the WS-53K SAE bracket-depth profile, which is bimodal (near-equal peaks at passes 2 and 8) rather than single-peaked, spanning a $-2$ to $+4$ pass range relative to WS-206K. The overall picture is rate, not target: WS-53K reaches the same milestones on a slightly delayed schedule.

\begin{table}[H]
\centering
\caption{Capacity comparison across diagnostics. ``Shift'' is WS-53K minus WS-206K peak pass (positive = WS-53K peaks later). Probing and validity peaks are 3-seed means; SAE peaks are single-seed estimates; $\pm 1$-pass shifts are within probe noise.}
\label{tab:capacity}
\small
\begin{tabular}{llccc}
\toprule
Method & Measure & WS-53K & WS-206K & Shift \\
\midrule
Validity        & Per-pass validity peak    & pass 8      & pass 7       & $+$1 \\
Probing         & Bracket depth peak        & pass 3      & pass 3       & 0    \\
Probing         & Ring state peak           & pass 5      & pass 4       & $+$1 \\
SAE             & Bracket detector peak     & pass 2      & pass 1       & $+$1     \\
SAE             & Bracket depth peak        & passes 2, 8 & pass 4       & $-2$ / $+4$ \\
SAE             & Ring state peak           & pass 5      & pass 4       & $+$1     \\
\bottomrule
\end{tabular}
\end{table}

\section{Distillation ablation}
\label{app:kd_ablation}

An earlier version of this sweep used a label-smoothed teacher; smoothing made the soft-target signal nearly uniform and looked like evidence that KD helps. We re-ran the full sweep with a non-smoothed teacher (GPT-467K-mod; $\tau=2.0$), and the apparent benefit disappeared: every mixture of soft targets underperforms hard-only (Table~\ref{tab:kd_ablation}). The curriculum and curriculum+KD rows disambiguate the two effects: the scheduled data order alone gets 96.7\% validity, and adding soft targets on top drops it to 90.3\%, so the residual damage comes from soft targets themselves rather than from the training schedule.

\begin{table}[H]
\centering
\caption{Distillation ablation (WS-206K, 1 seed except curriculum+KD which is 3 seeds). Hard-only and curriculum rows use the distillation pipeline and should be compared within this table, not against Table~\ref{tab:pareto}.}
\label{tab:kd_ablation}
\small
\begin{tabular}{lccc}
\toprule
Condition & Alpha & Validity (\%) & FCD \\
\midrule
Hard-only & 1.0 & 98.10 & 2.23 \\
Curriculum & n/a & 96.70 & 2.34 \\
Mixed & 0.5 & 91.90 & 2.31 \\
Curriculum + KD & 0.5 & 90.33 $\pm$ 0.49 & 2.45 \\
Soft-only & 0.0 & 89.30 & 2.21 \\
\bottomrule
\end{tabular}
\end{table}

\section{Teacher quality}
\label{app:teacher}

Teacher validity spans ${\sim}1$ percentage point across the three candidates (GPT-3.2M 99.4\%, GPT-467K-mod 98.5\%, WS-206K self-distill 98.5\%). Distillation from the two external teachers yields student validities within 0.5 points (89.9\% and 90.2\%) despite one teacher having 7 times more parameters than the other, while self-distillation reaches 90.4\% (Table~\ref{tab:teacher}). All three students remain ${\sim}8$ points below the no-KD WS-206K baseline of 98.0\%, so teacher quality is not the binding constraint on the KD degradation. Results are single-seed.

\begin{table}[H]
\centering
\caption{Teacher quality (student is WS-206K, 1 seed).}
\label{tab:teacher}
\small
\begin{tabular}{lrcc}
\toprule
Teacher & Params & Student validity (\%) & FCD \\
\midrule
GPT-3.2M        & 3.2M & 89.9 & 2.18 \\
GPT-467K-mod    & 467K & 90.2 & 2.26 \\
WS-206K (self)  & 206K & 90.4 & 2.25 \\
\bottomrule
\end{tabular}
\end{table}

\section{Direct Preference Optimization}
\label{app:dpo}

We apply DPO to two weight-shared bases (WS-206K, WS-82K) and one curriculum-distilled WS-206K variant, using the settings in Appendix~\ref{app:configs}. Across all three bases, validity changes by less than 2 percentage points and FCD shifts by at most 0.16 (Table~\ref{tab:dpo}). All results are single-seed.

\begin{table}[H]
\centering
\caption{DPO before/after results (1 seed each).}
\label{tab:dpo}
\small
\begin{tabular}{lcccc}
\toprule
Base & Validity pre (\%) & Validity post (\%) & FCD pre & FCD post \\
\midrule
WS-206K                          & 98.5 & 97.6 & 2.52 & 2.58 \\
WS-82K                           & 96.7 & 96.5 & 2.58 & 2.74 \\
WS-206K (curriculum-distilled)   & 96.7 & 98.3 & 2.29 & 2.42 \\
\bottomrule
\end{tabular}
\end{table}

\end{document}